\documentclass{article}

\PassOptionsToPackage{numbers, compress}{natbib}

\usepackage[preprint]{neurips_2020}

\usepackage[utf8]{inputenc} 
\usepackage[T1]{fontenc}    
\usepackage{hyperref}       
\usepackage{url}            
\usepackage{booktabs}       
\usepackage{amsfonts}       
\usepackage{nicefrac}       
\usepackage{microtype}      

\usepackage{graphicx}
\usepackage{commath}
\usepackage{multirow}
\usepackage{subfig}
\usepackage{mathtools}

\title{Embedding Differentiable Sparsity into Deep Neural Network}

\author{%
  Yongjin Lee \\
  ETRI\\
  Daejeon, Korea \\
  \texttt{solarone@etri.re.kr} \\
}

\begin{document}

\maketitle

\begin{abstract}

In this paper, we propose embedding sparsity into the structure of deep neural networks, where
model parameters can be exactly zero during training with the stochastic gradient descent.
Thus, it can learn the sparsified structure and the weights of networks simultaneously.
The proposed approach can learn structured as well as unstructured sparsity.



\end{abstract}

\section{Introduction}
Deep neural networks have made great success in various recognition and prediction problems,
but the size and the required computing resource have been overgrown as well.
To reduce its size as well as computation time for inference,
several approaches have been attempted.
Among them, pruning has long been adapted~\cite{Mozer1988,LeCun1989OBD,Hassibi1993,Liu2015,Han2015,Han2016,Dong2017}.
However, it typically requires a pre-trained model and needs to go through several steps:
selects unimportant parameters of a pre-trained model, deletes them and then, retrains the slimmed model,
and may repeat the whole process multiple times.

Another most recognized approach is the sparse regularization with $l_1$-norm,
which shrinks redundant parameters to zero during training~\cite{Tibshirani1996}, and thus does not require a pre-trained model.
However, since it acts on an individual parameter, it often produces unstructured irregular models
and thus, diminishes the benefit of computation on parallel hardware such as GPUs~\cite{Wen2016}.
In order to obtain regular sparse structures,
the regularization with $l_{2}$-norm~\cite{Yuan2007} was adopted on a group of parameters,
where a group was defined as a set of parameters on the same filter, neuron, layer or building block,
so that all parameters under the same group were either retained or zeroed-out together~\cite{Alvarez2016,Wen2016,Yoon2017}.
The optimization of the regularized objective is performed with proximal operation~\cite{Yuan2007,Parikh2014}.
The proximal operation is involved with soft-thresholding which consists of weight-decaying and thresholding operations
and it is carried out as a separate step from the gradient descent-based optimization for a prediction loss.
Therefore, it interrupts an end-to-end
training with the stochastic gradient descent and
it does not necessarily well balance the prediction loss and the model complexity.

In this work, we propose a differentiable approach in an end-to-end manner.
Our method allows model parameters to be exactly zero
during training with the stochastic gradient descent
and thus it does not require both a proximal operator or a parameter selection stage
in order to sparsify a model.
Since it can learn the sparsified structure and the weights of networks simultaneously by optimizing an objective function,
it abstracts and simplifies the whole learning process.
Another advantage of the proposed method is that it can be applied on a group of parameters,
and thus it can produce a structured model.

\section{Related Work}
The proposed approach is inspired by the group regularization with proximal operators~\cite{Zhou2010,Alvarez2016,Wen2016,Yoon2017}
and a differentiable sparsification approach~\cite{Lee2019sparsity}.
We briefly review the two approaches in order to show how we were motivated.

\subsection{Group Sparse Regularization}
The group regularization with $l_{2,1}$-norm enforces the sparsity in a group level.
A group is defined as a set of parameters on the same filter, neuron or layer,
and all parameters under the same group are either retained or zeroed-out together.
The group sparsity was successfully applied to automatically determine
the number of neurons and layers~\cite{Alvarez2016,Wen2016}.
The regularized objective function with $l_{2,1}$-norm~\cite{Yuan2007} is written as
\begin{equation}
  \mathcal{L} \left(  D, W\right) + \lambda \mathcal{R} \left(W \right),
\end{equation}
where $\mathcal{L}$ and $\mathcal{R}$ denote a prediction loss and a regularization term respectively,
and $D$ is a set of training data, $W$ a set of model parameters, and
$\lambda$ controls the trade-off between a prediction loss and a model complexity.
The regularization term is written as
\begin{equation}\label{eqn:group_sparsity}
\mathcal{R}\left( W\right) = \sum_{g}\norm{\textbf{w}_g}_{2} = \sum_{g}\sqrt{\sum_{i}w_{g,i}^{2} },
\end{equation}
where $W=\{ \textbf{w}_g\}$ and $\textbf{w}_g$ represents a group of model parameters.
In order to optimize the regularization term, parameter updating is performed with proximal operation~\cite{Yuan2007,Parikh2014},
\begin{equation}\label{eqn:group_prox}
 w_{g,i} \leftarrow \left( \frac{\norm{\textbf{w}_g}_2 - \eta\lambda}{\norm{\textbf{w}_g}_2} \right)_{+} w_{g,i},
\end{equation}
where $\leftarrow$ denotes an assignment operator and $\eta$ is a learning rate.

Another notable group regularization is exclusive lasso with $l_{1,2}$-norm~\cite{Zhou2010,Yoon2017}.
Rather than either retaining or removing an entire group altogether,
it was employed to promote sparsity within a group. 
The regularization term is written as
\begin{equation}\label{eqn:exc_sparsity}
\mathcal{R}\left( W\right) = \frac{1}{2} \sum_{g}\norm{\textbf{w}_g}_{1}^{2} = \frac{1}{2} \sum_{g}\left(\sum_{i}\abs{w_{g,i}} \right)^{2}.
\end{equation}
To optimize the regularization term, learning is performed with the following proximal operator,
\begin{equation}\label{eqn:exc_prox}
w_{g,i} \leftarrow sign\left(w_{g,i}\right) \left(\abs{w_{g,i}} - \eta \lambda \norm{\textbf{w}_g}_{1}\right)_{+}.
\end{equation}

The proximal operators consist of
weight decaying and thresholding steps and they are performed at every mini-batch or epoch in a sperate
step after the optimization of a prediction loss~~\cite{Alvarez2016,Wen2016,Yuan2007}. Thus, the parameter updating with the proximal
gradient descent can be seen as an additional model discretization or a pruning step. Moreover, because it
is carried out as a separate step from the optimization for a prediction loss,
it does not necessarily well balance the model complexity with the prediction loss.


\subsection{Differentiable Sparsification}
The differentiable approach~\cite{Lee2019sparsity} embeds trainable architecture parameters into the structure of neural networks.
By creating competition between the parameters and driving them to zero, they remove redundant or unnecessary components.
For example, suppose that a neural network is given as a modularized form,
\begin{equation}\label{eqn:base}
  \textbf{y}\left( \textbf{x}\right) = \sum_{i=1}^{n}{a_i  \textbf{f}_i\left(\textbf{x}; \textbf{w}_i\right)},
\end{equation}
where $\textbf{x}$ denotes an input to a module, $\textbf{w}_i$ model parameters for component $\textbf{f}_i$
and $a_{i}$ an architecture parameter.
Model parameters $\textbf{w}_i$ denote ordinary parameters such as a filter in a convolutional layer or
a weight in a fully connected layer.
The value of $a_i$ represents the importance of component $i$.
Enforcing $a_i$ to be zero amounts to removing component $\textbf{f}_i$ or zeroing-out whole $\textbf{w}_i$.
Thus, by creating the competition between elements of $a$ and driving some of them to be zero,
unnecessary components can be removed.

In order to allow the elements of $a$ to be zero and set up the competition between them,
the differentiable approach~\cite{Lee2019sparsity} parameterizes architecture parameters as follows:
\begin{equation} \label{eqn:gamma}
 \gamma_{i} = \exp\left(\alpha_i\right)
\end{equation}
\begin{equation} \label{eqn:thresholding}
  \tilde{\gamma}_{i} = \left( \gamma_{i} - \sigma\left( \beta \right) \norm{\gamma}_{1} \right)_{+}
\end{equation}
\begin{equation}\label{eqn:softmax}
    a_{i} = \frac{\tilde{\gamma}_i}{\sum_{j=1}^{n}{\tilde{\gamma}_j}} ,
\end{equation}
where $\alpha_i$ and $\beta$ are unconstrained trainable parameters, $\sigma(\cdot)$ denotes a sigmoid function and $\left(\cdot\right)_{+}$ represents $relu(\cdot) = \max(\cdot, 0)$.
It can be easily verified that $a_i$ is allowed to be zero and it is also differentiable in the view of modern deep learning.
In addition, they employ $p$-norm with $p < 1$ in order to further encourage the sparsity of $a$ or
the competition between its elements.
\begin{equation}
    \mathcal{R} \left(a \right) = \left(\sum_{i=1}^{n} {\abs{a_{i}}^{p}} \right) ^{\frac{1}{p}}. 
\end{equation}
The proximal operator of Eq.~(\ref{eqn:exc_prox}) is reduced to the form of Eq.~(\ref{eqn:thresholding}),
when the model parameters are non-negative.
Although their forms are similar to each other, they have completely different meanings.
The proximal operator is a learning rule whereas Eq.~(\ref{eqn:thresholding}) is the parameterized form of architecture parameters,
which is the part of a neural network.
Inspired by these two approaches, we directly embeds the learning rule into the structure of deep neural networks
by re-parameterizing model parameters.

\section{Proposed Approach}
In this section, we derive two methods for embedding sparsity into a deep neural network.
One is for structure sparsity and the other one is for unstructured sparsity.
We simply distinguish two approaches in order to show how we are motivated and where they are derived from.
With proper regularizers, the embedding method for unstructured sparsity can be used to induce structured sparsity as well.

\subsection{Structured Sparsity}
Motivated by the proximal operator of Eq.~\ref{eqn:group_prox}
and the threshold operation of Eq.~\ref{eqn:thresholding},
we directly embed the proximal operator into a deep neural network 
without resorting to architecture parameters. 
An original variable $w_{g,i}$ is re-parameterized as
\begin{equation}\label{eqn:group_param}
 \tilde{w}_{g,i} =  \left( \frac{\norm{\textbf{w}_g}_2 - \exp\left(\beta_g\right)}{\norm{\textbf{w}_g}_2} \right)_{+} w_{g,i},
\end{equation}
where $\left(\cdot\right)_{+}=\max(\cdot, 0)$, and 
$\tilde{w}_{g,i}$ is used instead as an ordinary parameter such as convolutional filters.
As in the proximal operator,
if the magnitude of a group is less than $\exp(\beta_g)$,
all parameters within the same group are zeroed-out.
Note that $\beta_g$ is not a constant, but it is a trainable parameter and is adjusted
by considering the trade-off between a prediction loss and a regularization term through training.
As notified by~\cite{Lee2019sparsity}, considering the support of \emph{relu} as a built-in
differentiable function in a modern deep learning tool,
it should not matter with the stochastic gradient descent-based learning.
Thus, our approach can simultaneously optimize the prediction accuracy and
the model complexity using the stochastic gradient descent.

We want to eventually drive model parameters to be zero, and thus
$\norm{\textbf{w}_{g}}_2$ in the denominator can cause numerical problems
when $\textbf{w}_{g}=0$. Since $\textbf{w}_g=0$ implies $w_{g,i}=0$, we can set
$\tilde{w}_{g,i}$ to zero when $\textbf{w}_{g}=0$, or we simply add a small number to the denominator.
Also, we can reformulate it with a scaling factor:
\begin{equation}\label{eqn:group_param_scale}
 \tilde{w}_{g,i} =  \left( \sigma\left(\alpha_g\right) \norm{\textbf{w}_g}_2 - \sigma\left(\beta_g\right) \right)_{+} w_{g,i},
\end{equation}
where $\alpha_g$ denotes a learnable scale parameter.


\subsection{Unstructured Sparsity}
Similarly as in the previous section, unstructured sparsity can be embedded by re-parameterizing an original variable $w_{g,i}$ as
\begin{equation}\label{eqn:exc_param}
\tilde{w}_{g,i} = sign\left(w_{g,i}\right) \left(\abs{w_{g,i}} - \sigma\left(\beta_g\right) \norm{\textbf{w}_g}_{1}\right)_{+},
\end{equation}
which are motivated by Eq.~(\ref{eqn:group_prox}) and~(\ref{eqn:thresholding}).
An individual parameter is zeroed-out according to the relative magnitude with its group
whereas Eq.~(\ref{eqn:group_param}) tends to remove an entire group altogether.
However, it does not mean that Eq.~(\ref{eqn:exc_param}) cannot induce structured sparsity.
A group regularizer such as $l_{2,1}$-norm of Eq.~\ref{eqn:group_sparsity} derives parameters within the same group
to have similar values and thus it can remove all parameters by rasing the threshold.

The gradient of the $sign$ function is zero almost everywhere, but it does not cause a problem for a modern deep learning tool.
The equation can be rewritten as
\[
    \tilde{w}_{g,i} =
\begin{dcases}
    \left(w_{g,i} + \sigma\left(\beta_g\right) \norm{\textbf{w}_g}_{1}\right)_{-} & \text{if } w_{g,i} < 0\\
    \left(w_{g,i} - \sigma\left(\beta_g\right) \norm{\textbf{w}_g}_{1}\right)_{+} & \text{otherwise},
\end{dcases}
\]
where $\left(\cdot\right)_{-}=\min(\cdot, 0)$.
We can handle it separately according to whether its value is negative or not.
For example, the conditional statement can be implemented using \emph{tf.cond} or \emph{tf.where} of TensorFlow~\cite{tensorflow}.

\subsection{Regularized Objective Function}

In the proposed approach, a regularized objective function can be written as
\begin{equation}
  \mathcal{L} \left(D, W, \beta\right) + \lambda \mathcal{R} \left(\tilde{W} \right),
\end{equation}
where $D$ is a set of training data, $W$ and $\tilde{W}$ denote a set of original and reformulated model parameters respectively,
and $\beta$ represents a set of threshold parameters.
In usual, a regularization is applied on free parameters, i.e, $W$, but
it is more appropriate to apply the regularization on $\tilde{W}$
since $\tilde{W}$ is the function of $\beta$:
the threshold parameter can receive the learning signal directly from the regularization term
and thus it can better learn how to balance the model complexity with the prediction loss.

Although the embedding forms of Eq.~(\ref{eqn:group_param}) and (\ref{eqn:exc_param}) are
derived from the $l_{2,1}$-and $l_{1,2}$-norm (Eq.~(\ref{eqn:group_sparsity}) and~(\ref{eqn:exc_sparsity})) respectively,
they do not need to be paired with their origins.
They are agnostic to any regularizers.
We can adopt $l_2$ regularizer, $\norm{\tilde{\textbf{w}}_g}^2_2$,
or even $p$-norm with $p < 1$ as in~\cite{Lee2019sparsity},
\begin{equation}
    \mathcal{R} \left(\tilde{W} \right) = \sum_{g}\norm{\tilde{\textbf{w}}_g}_{p}
    = \sum_{g} \left(\sum_{i} {\abs{\tilde{w}_{g,i}}^{p}} \right) ^{\frac{1}{p}}.
\end{equation}
The $p$-norm is well-known for its nature of inducing sparsity, but it is rarely used in practice
because it is not convex and the efficient optimization that makes parameters exactly zero is not known.
Thus, $l_1$-norm is widely used instead as a convex surrogate even if the $p$-norm is more ideal for inducing sparsity.
In our approach, however, we can employ any kinds of regularizers
as long as they are differentiable almost everywhere.

\subsection{Coarse Gradient for Threshold Operation}
If the magnitude of a parameter or a group is less than a threshold,
it is zeroed-out by \emph{relu} and it does not receive a learning signal
since the gradient of \emph{relu} is zero on the negative domain.
However, it does not necessary mean that it permanently dies
once its magnitude is less than a threshold.
It still has a chance to recover
because a threshold is adjustable and
the magnitude of a parameter can be increased by receiving the learning signal from a prediction loss.
Nevertheless, in order to make sure that dropped parameters can receive learning signals and have more chances to recover,
we can approximate the gradient of the thresholding function.
Previously, \citet{Xiao2019} approximated the gradient of a step function
using that of \emph{leaky relu} or \emph{soft plus} and
\citet{Lee2019sparsity} used \emph{elu} for \emph{relu}.
We follow the work of \citet{Lee2019sparsity} in order to improve the recoverability:
\emph{relu} is used in the forward pass but \emph{elu} is used in the backward pass.
As notified by~\citet{Lee2019sparsity}, this heuristics can be easily implemented
using modern deep learning tools and it does not interrupt an end-to-end learning with the stochastic gradient descent.

\subsection{Gradual Sparsity Regularization}\label{session:gradual_reg}

In the early stage of training, it is difficult to figure out which parameters are necessary and which ones are not
because they are randomly initialized.
Gradual scheduling $\lambda$ can help to prevent the accidental early dropping.
In addition, it changes the structure of the neural network smoothly
and helps the learning process to be more stable.
Inspired by the gradual pruning of~\cite{Zhu2017}, we increase the value of $\lambda$
from an initial  value $\lambda_i$ to a final value $\lambda_f$ over a span of $n$ epochs.
Starting at epoch $t_0$, we increase the value at every epoch:
\[
\lambda_t = \lambda_f + \left( \lambda_i - \lambda_f \right) \left( 1 - \frac{t-t_0}{n} \right)^3,
\] where $t$ denotes an epoch index.


\section{Conclusion}

We have proposed a gradient-based sparsification method that
can learn the sparsified structure and the weights of networks simultaneously.
Our proposed method can be applied on structured as well as unstructured sparsity.


%
%

\begin{ack}
This research was supported by the National Research Council of Science \& Technology (NST) grant by the Korea government (MSIP) (No. CRC-15-05-ETRI).
\end{ack}

\medskip

\small

\bibliography{reference}
\bibliographystyle{plainnat}

\end{document}